\pdfoutput=1
\documentclass[11pt]{article}

\usepackage[final]{acl}

\usepackage{times}
\usepackage{latexsym}

\usepackage[T1]{fontenc}
\usepackage[utf8]{inputenc}

\usepackage{microtype}

\usepackage{inconsolata}

\usepackage{graphicx}
\usepackage{hyperref}       % hyperlinks
\usepackage{url}            % simple URL typesetting
\usepackage{booktabs}       % professional-quality tables
\usepackage{amsfonts}       % blackboard math symbols
\usepackage{nicefrac}       % compact symbols for 1/2, etc.
\usepackage{microtype}      % microtypography
\usepackage{xcolor}         % colors
\usepackage{booktabs}
\usepackage[normalem]{ulem}
\usepackage{colortbl}
\usepackage{multirow}
\usepackage{amsmath}
\useunder{\uline}{\ul}{}
\definecolor{cGrey}{HTML}{F3F7F2}

\usepackage{arydshln}

\title{Self-Training Large Language and Vision Assistant for Medical Question-Answering}
\author{
 \textbf{Guohao Sun\textsuperscript{1}},
 \textbf{Can Qin\textsuperscript{2}},
\textbf{Huazhu Fu\textsuperscript{3}},
 \textbf{Linwei Wang\textsuperscript{1}},
 \textbf{Zhiqiang Tao\textsuperscript{1}},
\\
 \textsuperscript{1}Rochester Institute of Technology,
 \textsuperscript{2}Salesforce AI Research,\\
 \textsuperscript{3}Institute of High Performance, Computing, Agency for Science, Technology and Research
\\
 \texttt{\{gs4288, linwei.wang, zhiqiang.tao\}@rit.edu}\\
 \texttt{cqin@salesforce.com}, \texttt{hzfu@ieee.org}
}

\begin{document}
\maketitle
\begin{abstract}
Large Vision-Language Models (LVLMs) have shown significant potential in assisting medical diagnosis by leveraging extensive biomedical datasets. However, the advancement of medical image understanding and reasoning critically depends on building high-quality visual instruction data, which is costly and labor-intensive to obtain, particularly in the medical domain. To mitigate this data-starving issue, we introduce \textbf{S}elf-\textbf{T}raining \textbf{L}arge \textbf{L}anguage \textbf{a}nd \textbf{V}ision \textbf{A}ssistant for \textbf{Med}icine (STLLaVA-Med). The proposed method is designed to train a policy model (an LVLM) capable of auto-generating medical visual instruction data to improve data efficiency, guided through Direct Preference Optimization (DPO). Specifically, a more powerful and larger LVLM (e.g., GPT-4o) is involved as a biomedical expert to oversee the DPO fine-tuning process on the auto-generated data, encouraging the policy model to align efficiently with human preferences.
We validate the efficacy and data efficiency of STLLaVA-Med across three major medical Visual Question Answering (VQA) benchmarks, demonstrating competitive zero-shot performance with the utilization of only 9\% of the medical data. Our implementation is available at \url{https://github.com/heliossun/STLLaVA-Med}.
\end{abstract}

%Initially, we fine-tune the policy model on meticulously curated biomedical instruction data to develop visual questioning and answering competencies. Subsequently, we engage biomedical experts to oversee a DPO fine-tuning process on the auto-generated medical data, ensuring the policy model aligns efficiently with human preferences.

\section{Introduction}
\begin{figure}[t]
  \centering
\includegraphics[width=\linewidth]{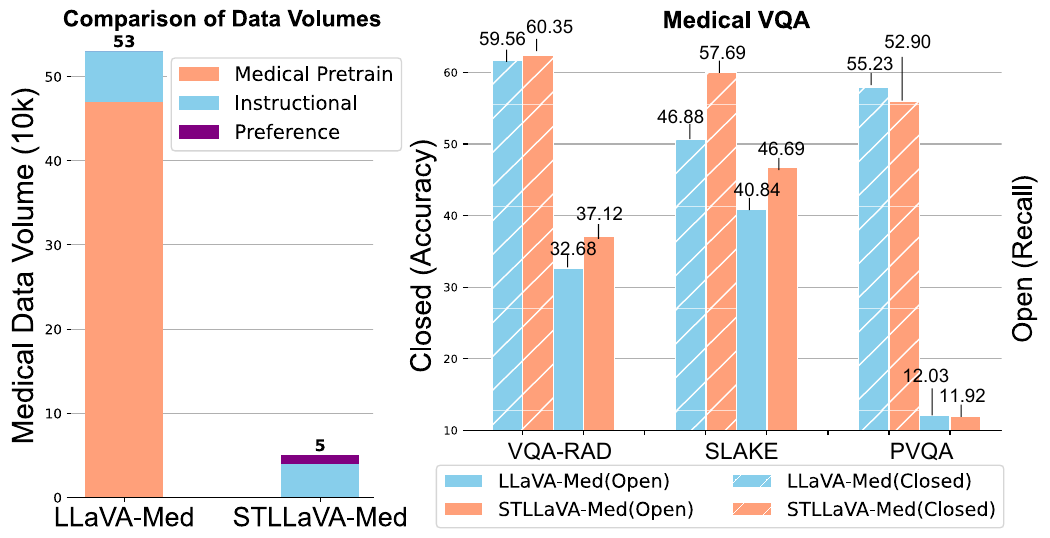}
  \vspace{-1.5em}
\caption{\label{fig:cover}\emph{Left:} Comparison of total medical data usage between LLaVA-Med (530K) and STLLaVA-Med (50k). \emph{Right:} Comparison results on three medical VQA datasets. STLLaVA-Med reports better/comparable performance, using much less medical training data.}
  \vspace{-1.5em}
\end{figure}

Large Vision-Language Models (LVLMs) have demonstrated impressive performance across a wide range of medical challenges~\cite{li2023llavamed, Moor2023MedFlamingoAM, Hu2024OmniMedVQAAN} by fine-tuning through biomedical visual instruction data. Similar to general LVLMs~\cite{llava1-5, Chen2023ShareGPT4VIL}, existing methods tailored for biomedical tasks primarily focus on collecting high-quality medical data to enhance task generalization and visual understanding. However, collecting medical data necessitates specialized expertise from physicians and raises privacy concerns, making the process both time-consuming and costly.
To address this data-starving issue, recent studies~\cite{li2023llavamed} have explored leveraging larger models/APIs (e.g., GPT-4~\cite{Achiam2023GPT4TR}) to generate medical data. Nevertheless, this kind of method does not fully resolve the high API costs~\cite{Deng2024EnhancingLV} associated with building instructional data and still requires large-scale pre-training data to align medical images and text (see Fig.~\ref{fig:cover} \emph{left}).

\begin{figure*}[t]
\centering
\includegraphics[width=\textwidth]{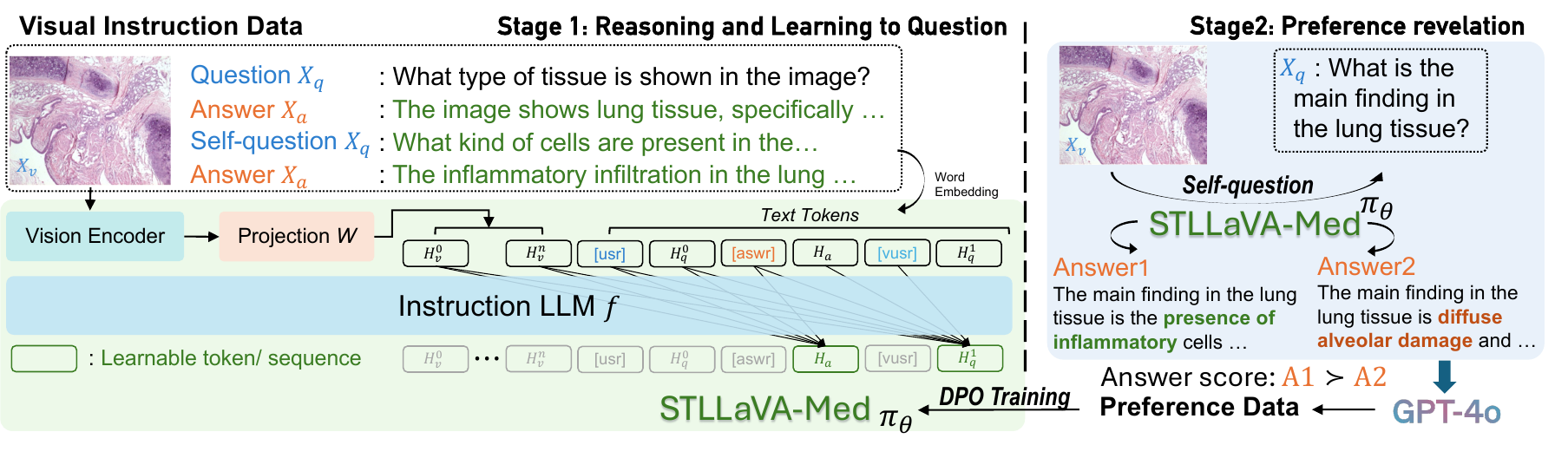}
\vspace{-2em}
\caption{ \label{fig:framework} Model architecture of STLLaVA-Med and self-training pipeline. \emph{Left}: stage 1 aiming to optimize the model $\pi_\theta$ improving medical image reasoning and learning to question. \emph{Right}: in stage 2, we first prompt $\pi_\theta$ to auto-generate preference data under the guidance of GPT-4o, then supervise $\pi_\theta$ for DPO fine-tuning. }
  \vspace{-1.5em}
\end{figure*}

To bridge the gap in medical data acquisition, we propose \textbf{S}elf-\textbf{T}raining \textbf{L}arge \textbf{L}anguage \textbf{a}nd \textbf{V}ision \textbf{A}ssistant for \textbf{Med}icine (STLLaVA-Med), a new training pipeline that enables LVLMs to automatically generate medical instruction data governed by Direct Preference Optimization (DPO)~\cite{rafailov2023direct}.
Different from previous self-training approaches~\cite{wang2023self, Deng2024EnhancingLV}, which generate answers for fixed/pre-defined questions (e.g., summarization and report), this work automatically generates open-ended questions and answers them, to enhance the diversity of self-training data and further improve medical image reasoning. 

Moreover, achieving precise control of the generated model response is also challenging due to its unsupervised nature~\cite{rafailov2023direct, Zhao2023BeyondHE, Azar2023AGT, Mehta2023SampleER}. Existing methods for gaining such steerability, such as reinforcement learning~\cite{Ouyang2022TrainingLM} and DPO~\cite{rafailov2023direct} from human feedback, mainly rely on collecting human labels to evaluate the relative quality of model generations and fine-tune the unsupervised LVLM to align with human preferences, which still burdens data collection in biomedical domains. To this end, the proposed STLLaVA-Med implements DPO by leveraging a larger LVLM with better general medical knowledge to supervise the policy model.

Overall, the proposed STLLaVA-Med realizes self-training in two stages --  \emph{1) reasoning and learning to question} and \emph{ 2) preference revelation}. In Stage 1, to enhance the model's reasoning and questioning skills, we incorporate questions within the visual instructional data as an additional learning objective following~\cite{Sun2024SQLLaVASF}. After the first-stage training, STLLaVA-Med can generate question-answer pairs automatically. In Stage 2, we leverage GPT-4o~\cite{2024GPT4o} as a medical expert to further supervise fine-tuning STLLaVA-Med through DPO, ensuring it adheres to our designed preferences (e.g., detail, relevance, and accuracy) on the auto-generated data. We summarize the contributions of this work as follows:
\begin{itemize}
    \item We propose a novel self-training approach for LVLMs that enhances medical reasoning skills with less medical data. Our approach improves the data efficiency of training LVLMs for specific domains.
    \item The proposed STLLaVA-Med enables the automatic construction of medical instructional data, supervised by a stronger and heavy LVLM (i.e., GPT-4o) and governed through DPO, which allows our LVLM to adhere to preferences in a self-training way. 
    %Direct Preference Optimization (DPO) ensures that STLLaVA-Med adheres to preferences during self-training.
    \item Experiments on three major medical VQA benchmarks demonstrate that our method achieves highly competitive zero-shot performance compared to existing methods yet utilizing only 9\% of the medical data.
\end{itemize}

% \ZT{DPO is the key to enable self-training and the main difference compared with self-questioning. Yet, it never appears in Abs and Intro. Please revise the Intro thoroughly.}

% \ZT{Contributions: 1) Introduce ST to medical LVLMs to enhance visual reasoning/understanding + data efficiency; 2) how we achieve self-training: gpt4o+ DPO; 3) Experiments.}

% \section{Related Work}
% \noindent\textbf{Large vision-language model for medical.}

% \noindent\textbf{Alignment fine-tuning.}
%  However, achieving precise control of their behavior is difficult due to the completely unsupervised nature of their training. Existing methods for gaining such steerability collect human labels of the relative quality of model generations and fine-tune the unsupervised LVLM to align with these preferences, often with reinforcement learning from human feedback (RLHF)~\cite{Ouyang2022TrainingLM} or Direct Preference Optimization (DPO and its variational)~\cite{rafailov2023direct, Dubois2023AlpacaFarmAS, Azar2023AGT,Mehta2023SampleER}. 

\section{STLLaVA-Med}
In this section, we introduce STLLaVA-Med (see Fig.~\ref{fig:framework}) given by our proposed two-stage self-training algorithm, which is designed to enhance the data efficiency when training an LVLM for medical tasks. Specifically, we optimize the LVLM -- a policy model -- in two stages sequentially. The policy model $\pi_\theta$ parameterized by $\theta$ first learns to automatically generate question-answer pairs for self-training, then utilizes DPO to control the prediction behavior precisely. 

\noindent\textbf{Stage1: Reasoning and learning to question.}\\
%We incorporate visual questioning into model training for better visual reasoning and enabling self-training. Intuitively, 
The main part of self-training is automatic question generation and answering. Specifically, we follow~\cite{Sun2024SQLLaVASF} by adding a special token \emph{<vusr>} and set the question-tokens to learnable, to jointly fine-tune $\pi_\theta$ for reasoning and questioning on visual instructional data $\mathcal{D}_{ft}$.

Given the visual instruction data $\mathcal{D}_{ft}$ = $\{ (X_v,X_c) \}_1^N$, where the conversation $X_c= \{ X_q^{(j)},X_a^{(j)}  \}^M_{j=1}$ consists of $M$ QA pairs, the text $X_c$, and the image $X_v$ are encoded to sequential embeddings as $H_c= \{ H_q^{(j)},H_a^{(j)}  \}^M_{j=1}$ and $H_v$ by word embedding and vision encoder. We minimize the negative log-likelihood loss for the \emph{vq: visual questioning} and \emph{qa: answering} as the following:
 \begin{equation*}
\setlength{\abovedisplayskip}{1pt}
    \setlength{\belowdisplayskip}{1pt}
    {\small\begin{aligned}
    &\mathcal{L}_{vq}=\sum_{v, c\in \mathcal{D}_{ft}}^{}-\mathrm{log} \pi_\theta(H_q^{(j+1)} \mid H_v,H_{c}^{(1:j)}), \\
    &{\mathcal{L}_{qa}=\sum_{v, c\in \mathcal{D}_{ft}}^{}-\mathrm{log}\pi_\theta(H_a^{(j+1)} \mid H_v,H_{c}^{(1:j)},H_q^{(j+1)})},
\end{aligned}}
\end{equation*}
where $j\in\left\{ 1,\cdots ,M \right\}$ indicates the index of question or answer within the conversational data $H_c= \{ H_q^{(j)},H_a^{(j)}  \}^M_{j=1}$.
To mitigate the heavy computational overhead, we fine-tune LoRA~\cite{hu2021lora} in both the vision encoder and LLM. Thus, the learnable parameters $\theta$ of the policy model $\pi_\theta$ during fine-tuning represent a combination of all the parameters of LLM-LoRA, ViT-LoRA, and the vision-to-language projector. In addition, we skip the vision-language alignment on medical image-text pairs for data efficiency by loading the pre-trained weights~\cite{Sun2024SQLLaVASF} fine-tuned on general-purpose visual instructional data.
After training with these objective functions, our model can raise and answer questions for a medical image, enabling automatic QA generation in the following stage.

\noindent\textbf{Stage2: Preference revelation.}\\
We apply DPO~\cite{rafailov2023direct} to fine-tune the unsupervised LVLM to align with pre-defined preferences. Unlike previous works~\cite{rafailov2023direct, Zhao2023BeyondHE}, we employ $\pi_\theta$ to automatically generate a preference dataset $D_{pref}= \{ (X_v, X_q,X_{a_w},X_{a_l}) \}_1^N $. Specifically, we prompt $\pi_\theta$ to generate an image-related question $X_q$ and two different answers $X_{a}$, which are labeled as \emph{win} $X_{a_w}$ and \emph{loss} $X_{a_l}$ answers by GPT-4o. For all the experiments, we first map $X_q$, $X_{a_w}$, $X_{a_l}$, and $X_v$ to $H_q$, $H_{a_w}$, $H_{a_l}$, and $H_v$ using the same word embedding and vision encoder in stage 1, and fine-tune $\pi_\theta$ through DPO by minimizing the following negative log-likelihood loss: 
\begin{equation}
\setlength{\abovedisplayskip}{1pt}
    \setlength{\belowdisplayskip}{1pt}
    {\small\begin{aligned}
    &\mathcal{L}_{DPO}(\pi_\theta;\pi_{ref})=-\mathbb{E}_{v, q, a_w, a_l \in \mathcal{D}_{pref}} [ \mathrm{log}\sigma (\beta \mathrm{log} \\
    &\frac{\pi_\theta(H_{a_w}\mid H_v,H_q)}{\pi_{ref}(H_{a_w}\mid H_v,H_q)})- \beta \mathrm{log}\frac{\pi_\theta(H_{a_l}\mid H_v,H_q)}{\pi_{ref}(H_{a_l}\mid H_v,H_q)}) ],
\end{aligned}}
\end{equation}
where $\beta$ is a parameter controlling the deviation from the base reference policy $\pi_{ref}$, which prevents the policy model from deviating too far from the distribution of correct generation, as well as maintaining the generation diversity and preventing mode-collapse to single high-reward answers~\cite{rafailov2023direct}. Notably, $\pi_\theta$ and $\pi_{ref}$ are initialized from the same weights at beginning, and only $\pi_\theta$ is optimized during training. 
In this way, we fit an implicit reward to precisely control the model generation by the pre-defined preference such as accuracy and detail.  
%The use of DPO objective has shown notable improvement in downstream medical tasks after stage1, confirming that supervised preference optimization enhances multi-modal reasoning ability.
% \begin{figure*}[t]
% \centering
% \includegraphics[width=0.9\textwidth]{figures/promptGPT.pdf}
% \vspace{-1em}
%   \caption{ \label{fig:prompt} }
% \end{figure*}

\section{Self-training Datasets}
Medical LVLMs~\cite{li2023llavamed, Zhang2023PMCVQAVI, Moor2023MedFlamingoAM} generally adopt pre-training on massive medical data, to realize medical image-text alignment. 
However, the proposed STLLaVA-Med does not involve such a medical corpus pre-training, providing new insights into data efficiency.
To fine-tune the LVLM for medical tasks, we utilize a filtered open-source medical instructional dataset Med-60k-IM~\cite{li2023llavamed} as $\mathcal{D}_{ft}$ due to image unavailability. Table~\ref{tab:data} provides the medical data statistics for training LLaVA-Med and STLLaVA-Med.
We show how to employ the policy model to auto-generate a preference dataset for DPO fine-tuning in the following process:\\
\noindent\textbf{Auto-generated Questions}.
We randomly sample 10k medical images from Med-60k-IM datasets and prompt $\pi_\theta$ to generate questions. \\
\noindent\textbf{GPT-4o guided preference data collection.} We prompt $\pi_\theta$ to predict two answers to each generated question. Specifically, to ensure the difference between answers, we set the temperature scaling to 1.2, $TopK=100$, and $TopP=0.95$, encouraging the model to generate more diverse and non-repetitive output. In previous research~\cite{rafailov2023direct}, the preference data were annotated by human annotators. In contrast, this work utilizes GPT-4o as a simulated expert since we observe its excellent biomedical performance~\cite{Yue2023MMMUAM} and the best downstream task performance in Table~\ref{tab:main}. We prompt GPT-4o (see Appendix~\ref{sec:prefdata} for prompt design) with all the information to label the answers with \emph{win} or \emph{loss}, treated as $X_{a_w}$ and $X_{a_l}$ within $\mathcal{D}_{pref}$.
% To be specific, Med-60k-IM are sampled from the five most common imaging modalities:
% CXR (chest X-ray), CT (computed tomography), MRI (magnetic resonance imaging), histopathology,
% and gross (i.e., macroscopic) pathology. 
% VQA-RAD contains QA pairs generated by clinicians, where the images are evenly distributed over the head, chest, and abdomen. Questions are categorized into 11 categories: abnormality, attribute, modality, organ system, color, counting, object/condition presence, size, plane, positional reasoning, etc. SLAKE is a Semantically-Labeled Knowledge-Enhanced dataset for medical VQA. The original dataset contains Chinese and English QA, but we only consider the English subset in our study. Besides, SLAKE includes richer modalities and covers more human body parts than the currently available dataset, including the brain, neck, chest, abdomen, and pelvic cavity. PathVQA is a dataset of pathology images. Each image has questions about multiple aspects, such as location, shape, color, appearance, etc. Overall, the questions of Gen-Med are categorized into two types, with several varieties: open-ended questions such as why, what, how, where, etc., and closed-ended questions with one-word answers (yes/no). 

\begin{table}[t]
\centering
\caption{\label{tab:data} Statistics of medical training data.}
\vspace{-1em}
\scalebox{0.9}{
\begin{tabular}{lccc}
\toprule
Method            & \#Images & \#QA-Pairs \\ \midrule
LLaVA-Med$_{pt}$             & 467710   &   467710   \\
LLaVA-Med$_{ft}$            &56708      &  164231     \\ \midrule
ours & 37452    & 108545     \\ \bottomrule
\end{tabular}}
\vspace{-1em}
\end{table}

\begin{table*}[t]
\centering
\caption{\label{tab:main} Comparison with other methods on three benchmarks. Open questions are evaluated by Recall and F1 score, and closed questions are evaluated by accuracy. All models are using 7B LLM. STLLaVA-Med w/o DPO is the ablated version of our final model. Notably, LLaVA-Med was trained on the original Med-60k-IM~\cite{li2023llavamed}, which has 20k more samples than the Med-IM we used in this work due to image unavailability.}
\vspace{-0.5em}
\scalebox{0.7}{
\begin{tabular}{ll|ccc|ccc|ccc}
\toprule
\multirow{2}{*}{Dataset} & \multirow{2}{*}{Method} & \multicolumn{3}{c|}{VQA-RAD} & \multicolumn{3}{c|}{SLAKE} & \multicolumn{3}{c}{PVQA} \\
 &  & Recall & F1 Score & Closed & Recall & F1 Score & Closed & Recall & F1 Score & Closed \\ \midrule
 & GPT-4o~\cite{2024GPT4o} & 51.60&9.23&63.97&59.06&8.90&71.63&24.14&3.29&75.97 \\
 \hdashline
\multirow{2}{*}{w/o Med-IM} & LLaVA-v1.5~\cite{llava1-5} & 23.63 & 9.53 & 50.74 & 35.23 & 8.84 & 52.16 & 11.85 & \textbf{2.73} & 52.76 \\
 & SQ-LLaVA~\cite{Sun2024SQLLaVASF} & 23.91 & 6.29 & 52.57 & 40.04 & 9.65 & 57.45 & 11.24 & 2.63 & 53.73 \\
 & Med-Flamingo~\cite{Moor2023MedFlamingoAM}&10.32&10.37&52.21&8.46&7.67&37.02&1.23&1.24&45.59 \\
 &PMC-VQA~\cite{Zhang2023PMCVQAVI}&6.26&5.68&41.54&7.29&6.92&33.89&1.02&1.01&40.10\\
 \midrule
\multirow{3}{*}{Med-IM} & LLaVA-Med~\cite{li2023llavamed} & 32.68 & 8.65 & 59.56 & 40.84 & 8.21 & 46.88 & \textbf{12.03} & 2.47 & \textbf{55.23} \\
 & STLLaVA-Med w/o DPO& 33.81 & 10.37 & 59.16 & 40.13 & 10.97 & 55.53 & 10.38 & 2.68 & 52.05 \\
 & STLLaVA-Med & \cellcolor{cGrey}\textbf{37.12} & \cellcolor{cGrey}\textbf{10.83} & \cellcolor{cGrey}\textbf{60.35} & \cellcolor{cGrey}\textbf{46.69} & \cellcolor{cGrey}\textbf{11.46} & \cellcolor{cGrey}\textbf{57.69} & \cellcolor{cGrey}11.92 & \cellcolor{cGrey}\textbf{2.72} & \cellcolor{cGrey}52.90 \\ \bottomrule
% \multirow{3}{*}{Gen-Med} & LLaVA-v1.5~\cite{llava1-5} & 43.44 & 36.41 & 70.59 & 52.76 & 46.98 & 64.18 & 35.91 & 35.47 & 91.15 \\
%  &STLLaVA-Med w/o DPO & 52.07 & 45.38 & 75.74 & 56.10 & 50.77 & 67.31 & 38.05 & 37.76 & 92.13 \\
%  & STLLaVA-Med & \cellcolor{cGrey}\textbf{52.60} & \cellcolor{cGrey}\textbf{45.92} & \cellcolor{cGrey}\textbf{76.10} & \cellcolor{cGrey}\textbf{57.37} & \cellcolor{cGrey}\textbf{50.84} & \cellcolor{cGrey}\textbf{67.31} & \cellcolor{cGrey}\textbf{38.30} & \cellcolor{cGrey}\textbf{38.00} & \cellcolor{cGrey}\textbf{92.13}
%  \\ \bottomrule
\end{tabular}}\vspace{-1em}
\end{table*}

\section{Experiments}
\subsection{Implementation}
We follow~\cite{Sun2024SQLLaVASF} to construct our model architecture, including the visual encoder, image projector, prototype extractor, and the instructional LLM.
Our proposed self-training pipeline involves two stages.
In stage 1, we continually fine-tune the policy model $\pi_\theta$, initialized from~\cite{Sun2024SQLLaVASF} on instructional data with global batch size as $128$. During training, we insert LoRA\cite{hu2021lora} with $rank = 128$ and $\alpha = 256$ into the language model (LLM-LoRA) and LoRA with $rank = 32$ and $\alpha = 64$ into the vision encoder (ViT-LoRA). We optimize the model using AdamW~\cite{adamw} optimizer for one epoch by setting the learning rate to $2\times10^{-4}$ for LoRA, and $5\times10^{-5}$ for the other layers. In stage 2, we fine-tune $\pi_\theta$ on the auto-generated preference dataset. Similar to stage 1, we utilize LoRA for light-weight training. We optimize the model using AdamW~\cite{adamw} optimizer for one epoch by setting the learning rate to $5\times10^{-6}$ for LoRA, and $2\times10^{-5}$ for the other layers. We train the model on 4 A100s for 10 hours. 
\begin{figure*}[h]
\centering
\includegraphics[width=0.95\textwidth]{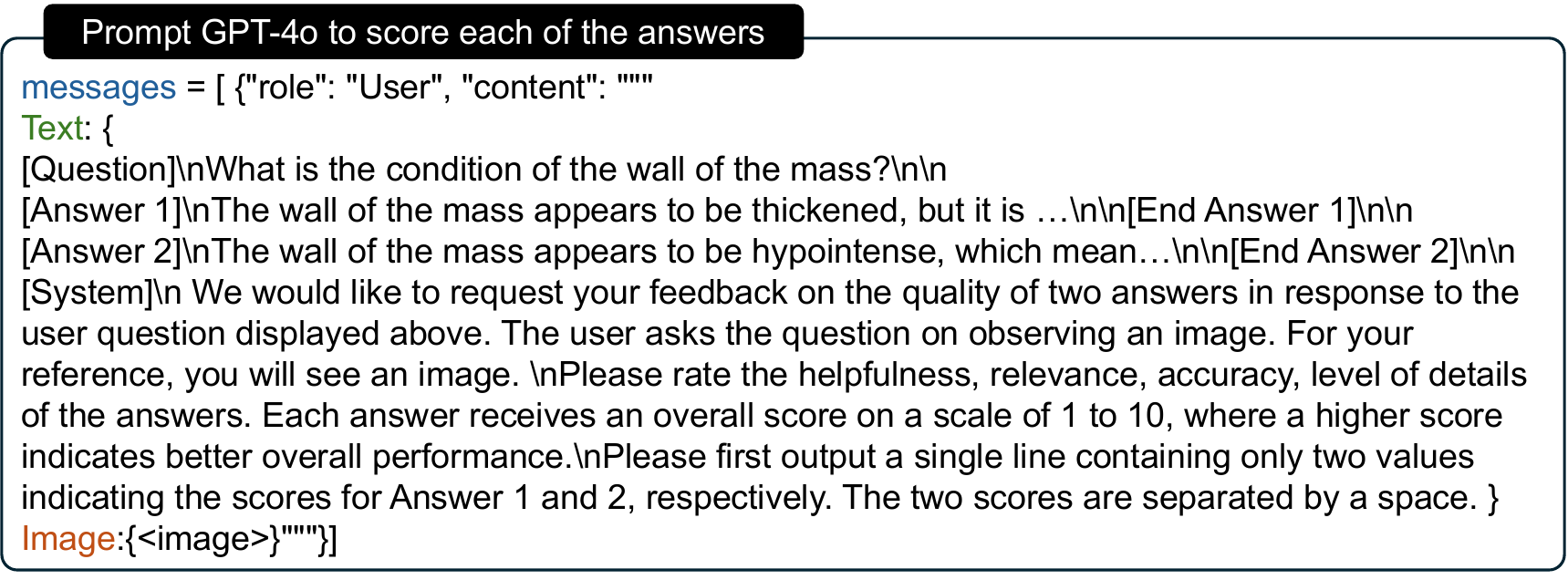}
\caption{ \label{fig:prompt} Prompt for GPT-4o to grade the answers generated by STLLaVA-Med from stage 1. The answer with the higher score will be designated as the winning response, while the other will be classified as rejected.}
\end{figure*}

\subsection{Preference Data Generation}\label{sec:prefdata}
In previous research~\cite{rafailov2023direct}, preference data were annotated by human annotators. In contrast, this work employs GPT-4o~\cite{2024GPT4o} as a simulated expert to classify the answers generated by STLLaVA-Med. As shown in Fig.~\ref{fig:prompt}, we provide the detailed prompt design for GPT-4o to label the answers with either \emph{win} or \emph{loss}. Due to its multi-modal understanding capabilities, GPT-4o can directly take images as input. In Appendix~\ref{sec:apenqualiresult}, we provide qualitative results about the preference data generated by STLLaVA-Med.

\subsection{Datasets and Metrics}
\noindent\textbf{Datasets.} We conduct experiments on the widely-used medical VQA benchmark dataset VQA-RAD~\cite{vqa-rad}, SLAKE~\cite{Liu2021SlakeAS}, and PVQA~\cite{He2020PathVQA3Q}. Specifically, VQA-RAD contains QA pairs generated by clinicians, where the images are evenly distributed over the head, chest, and abdomen. Questions are categorized into 11 categories: abnormality, attribute, modality, organ system, color, counting, etc. SLAKE is a Semantically-Labeled Knowledge-Enhanced dataset for medical VQA. The original dataset contains Chinese and English QA, but we only consider the English subset in our study. Besides, SLAKE includes richer modalities and covers more human body parts than the currently available dataset. PathVQA is a dataset of pathology images. Each image has questions about multiple aspects, such as location, shape, color, appearance, etc. Overall, the medical questions are categorized into two types: \textbf{open-ended} questions such as why, what, how, where, etc., and \textbf{closed-ended} questions with one-word answers (yes/no).

\noindent\textbf{Evaluation Metrics.} We report accuracy for closed questions. For open-ended questions, we compute recall as the proportion of correctly predicted words out of the reference sentence, and F1 score as a balance metric between recall and precision.

\subsection{Overall Performance}  
\begin{figure}[t]
  \begin{center}
    \includegraphics[width=\linewidth]{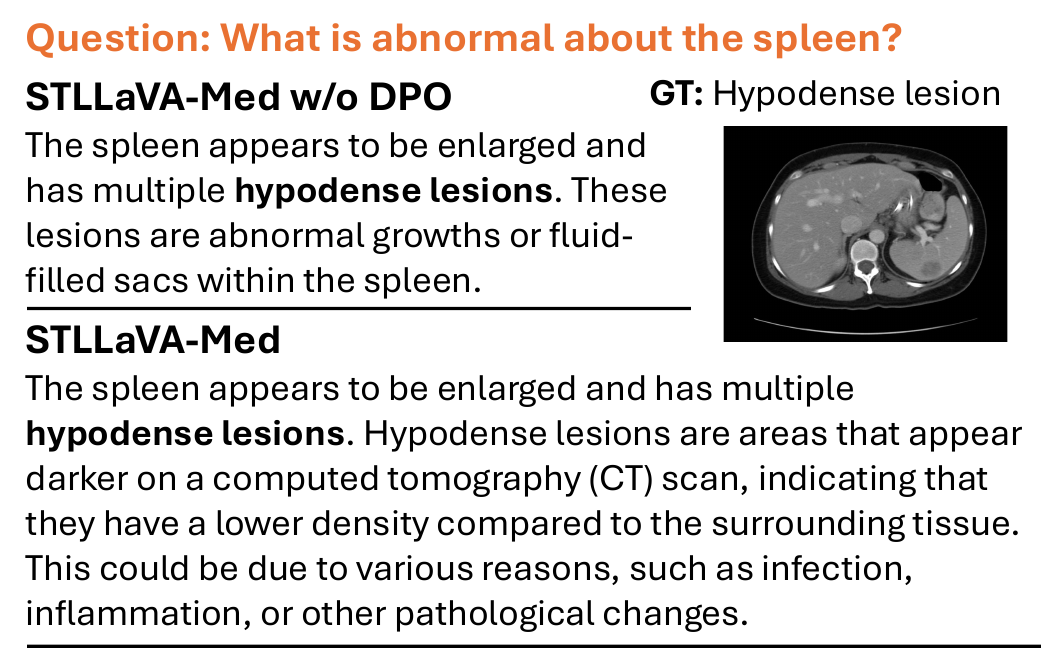}
  \end{center}
  \vspace{-1em}
  \caption{\label{fig:sample} Qualitative evaluation of methods \emph{w} and \emph{w/o} preference revelation. }
  \vspace{-2em}
\end{figure}

The evaluation results in Table~\ref{tab:main} are divided into three sections based on rows. We compile all the experiments locally with a single run. The first row indicates the upper bound of zero-shot medical performance; the next two rows and the last three rows reflect the performance of LVLMs trained without and with medical data. As shown in Table~\ref{tab:main}, even without pre-training on medical data, STLLaVA-Med achieves competitive performance with LLaVA-Med only after fine-tuned instructional data Med-IM. After DPO training, we observed a performance improvement over open-ended questions, demonstrating the effectiveness of supervised preference optimization. Also, in Fig.~\ref{fig:sample}, the answer of STLLaVA-Med has higher relevance, more detail, and accuracy than STLLaVA-Med w/o DPO, proving that self-training on the auto-generated preference dataset controls the model following pre-defined preference. See Appendix~\ref{sec:moreresults} for more results.
Moreover, we have several observations based on the results in Table~\ref{tab:main}.

1) Medical image-text alignment is unnecessary for LVLM-Med. As we observed, fine-tuning a pre-trained general-purpose LVLM (e.g., LLavA-V1.5~\cite{llava1-5}, SQ-LLaVA~\cite{Sun2024SQLLaVASF}) on a small set of medical instruction data achieves the same performance as fully training on medical data (LLaVA-Med). This suggests that general-purpose LVLM with strong vision-language alignment can be easily adapted to medical tasks after light-weight fine-tuning.

2) High-quality medical instruction data can further improve STLLaVA-Med, by enriching the auto-generated data's diversity, complexity, and professionalism.   

\section{Conclusion}
This work has proposed a self-training vision-language assistant for medicine (STLLaVA-Med), a novel training pipeline designed to enhance the data efficiency of training LVLMs for medical tasks. Our approach prompts the policy model to self-generate instruction-answer pairs and label them by a larger language model, such as GPT-4o, for preference optimization. This process aims to enhance the medical reasoning capabilities of a smaller vision-language model, reducing the reliance on extensively annotated medical data and alleviating human experts in the medical field. Experimental results on three benchmarks demonstrate that STLLaVA-Med achieves exceptional medical reasoning capabilities using medical data at a minimum level. We aspire for our work to inspire future research aimed at enhancing the efficiency of training LVLMs in broad medical domains.

\section{Limitations}
Although the proposed approach improves medical reasoning ability, the performance of self-training is highly dependent on the quality and relevance of the auto-generated medical instructional data. This indicates that we still need the instructional data from stage 1 training to cover a wider range of medical tasks and professional expertise, which may still be difficult to collect for some diseases or some types of medical images. In addition, GPT4o may become inevitable bias when annotating preference data. To address this, we may use another SOTA LMM, such as Gemini-pro~\cite{team2023gemini}, or include a medical expert in the loop to co-supervise the preference data collection process.
\section{Ethics Statements}
We conducted experiments and analysis on public datasets, PMC-15M, VQA-RAD, SLAKE, and PVQA, where all medical images and texts were de-identified, ensuring the privacy and confidentiality of patients. While our method reduces the need for extensive labeled datasets, its outputs are still machine-generated, requiring critical human oversight when used in clinical decision-making.
\section*{Acknowledgments}
This work is in part supported by NIH award R01NR018301.
\bibliography{main}

\appendix

\section{Related Work}
\subsection{Large Vision-Language Model}
As the field of Large Language Models (LLMs) and instruction tuning undergoes rapid advancements, the academic research community is increasingly focusing on integrating visual information into LLM frameworks to enhance vision-language modeling~\cite{clip,Li2022BLIPBL,10.1145/3589334.3645653,Wang_2024_CVPR,xue2024xgen}. This emerging research area has seen the development of various methodologies, building on the foundational work of CLIP~\cite{clip} and diverse LLM architectures such as Vicuna~\cite{vicuna}, Llama2~\cite{Touvron2023Llama2O}, and Qwen-VL~\cite{Bai2023QwenVLAV}.
Notably, LLaVA~\cite{llava} pioneered the integration of an LLM with a CLIP vision encoder to create a vision-language model, demonstrating significant capabilities in image-text dialogue tasks through strategies of pre-training alignment and targeted instruction tuning. Subsequent research has focused on refining visual instruction tuning by improving the quality and diversity of datasets used during the pre-training and fine-tuning phases.
Building upon these advancements, recent studies, including LLaVA-v1.5~\cite{llava1-5} and ShareGPT4V~\cite{Chen2023ShareGPT4VIL}, have achieved notable success in general vision-language comprehension, showcasing their ability to handle complex question-answering tasks. This progression underscores the importance of sophisticated data handling and model-tuning strategies in developing effective vision-language models.
\begin{figure}[t]
  \begin{center}
    \includegraphics[width=\linewidth]{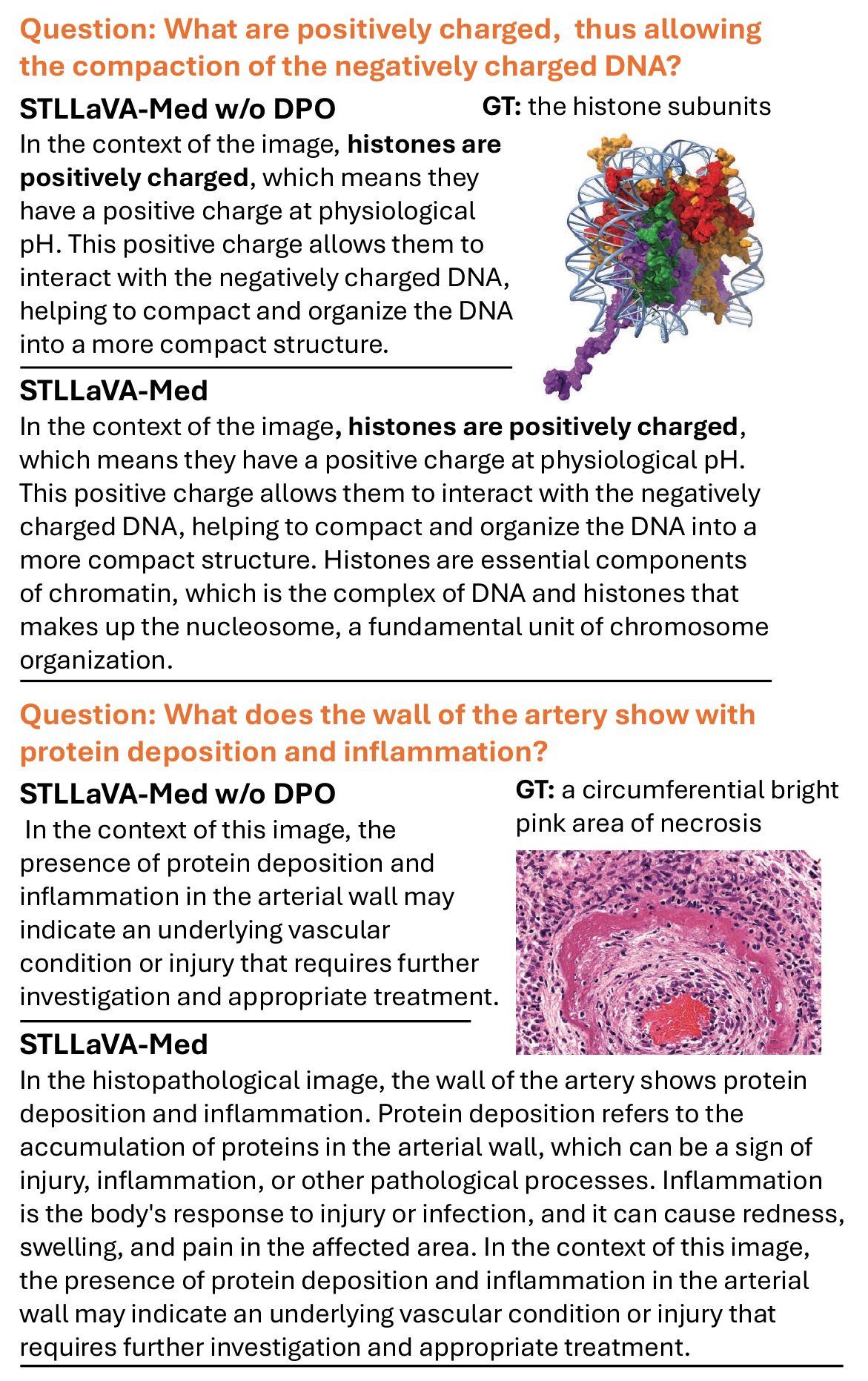}
  \end{center}\vspace{-1em}
  \caption{\label{fig:supsample} Qualitative evaluation of methods \emph{w} and \emph{w/o} preference revelation. }
\end{figure}

\begin{table}[h]
\centering
\caption{\label{tab:extradata} Medical data statistics of training.}
\scalebox{0.9}{
\begin{tabular}{lccc}
\toprule
Method            & \#Images & \#QA-Pairs \\ \midrule
Med-IM             & 56708      &  164231   \\
VQA-RAD             & 313   & 3064     \\
SLAKE           &  546   &  11934     \\ 
PVQA            & 37452    & 26034     \\ \bottomrule
\end{tabular}}
\end{table}
\begin{table*}[t]
\centering
\caption{\label{tab:ft} Comparison of fine-tuning performance on three benchmarks. Open questions are evaluated by Recall and F1 score, and closed questions are evaluated by accuracy. All models are using 7B LLM. STLLaVA-Med w/o DPO is the ablated version of our final model. }

\scalebox{0.73}{
\begin{tabular}{ll|ccc|ccc|ccc}
\toprule
\multirow{2}{*}{Dataset} & \multirow{2}{*}{Method} & \multicolumn{3}{c|}{VQA-RAD} & \multicolumn{3}{c|}{SLAKE} & \multicolumn{3}{c}{PVQA} \\
 &  & Recall & F1 Score & Closed & Recall & F1 Score & Closed & Recall & F1 Score & Closed \\ \midrule

\multirow{3}{*}{\begin{tabular}[c]{@{}l@{}}Med-IM+\\ VQA-RAD+\\ SLAKE+PVQA\end{tabular}} & LLaVA-v1.5~\cite{llava1-5} & 43.44 & 36.41 & 70.59 & 52.76 & 46.98 & 64.18 & 35.91 & 35.47 & 91.15 \\
 &STLLaVA-Med w/o DPO & 52.07 & 45.38 & 75.74 & 56.10 & 50.77 & 67.31 & 38.05 & 37.76 & 92.13 \\
 & STLLaVA-Med & \cellcolor{cGrey}\textbf{52.60} & \cellcolor{cGrey}\textbf{45.92} & \cellcolor{cGrey}\textbf{76.10} & \cellcolor{cGrey}\textbf{57.37} & \cellcolor{cGrey}\textbf{50.84} & \cellcolor{cGrey}\textbf{67.31} & \cellcolor{cGrey}\textbf{38.30} & \cellcolor{cGrey}\textbf{38.00} & \cellcolor{cGrey}\textbf{92.13}
 \\ \bottomrule
\end{tabular}}
\end{table*}

\noindent\textbf{Alignment fine-tuning.}
Following supervised fine-tuning (SFT), alignment fine-tuning has emerged as a key method to further enhance the performance of Large Language Models (LLMs) by aligning them with human preferences~\cite{Ouyang2022TrainingLM}. Initial approaches utilized on-policy reinforcement learning (RL) methods, such as proximal policy optimization (PPO)~\cite{Schulman2017ProximalPO}, to train a reward model based on preference data~\cite{Bai2022TrainingAH}.
The introduction of direct policy optimization (DPO)~\cite{rafailov2023direct,Dubois2023AlpacaFarmAS,Azar2023AGT,Mehta2023SampleER} has marked a significant shift towards direct learning from human preferences, bypassing the need for an explicit reward model. Another effective strategy is iterative preference fine-tuning, which repeatedly optimizes the model on newly generated preference pairs in successive iterations, thereby improving performance.
Despite extensive research on alignment fine-tuning for LLMs, the application of these techniques to Large Vision-Language Models (LVLMs) has been comparatively limited. Early attempts~\cite{Zhou2024AligningMI,Deng2024EnhancingLV} have focused on constructing preference datasets using human-labeled data or GPT-4 generations, followed by fine-tuning with a DPO loss.

\begin{figure*}[t]
\centering
\includegraphics[width=0.8\textwidth]{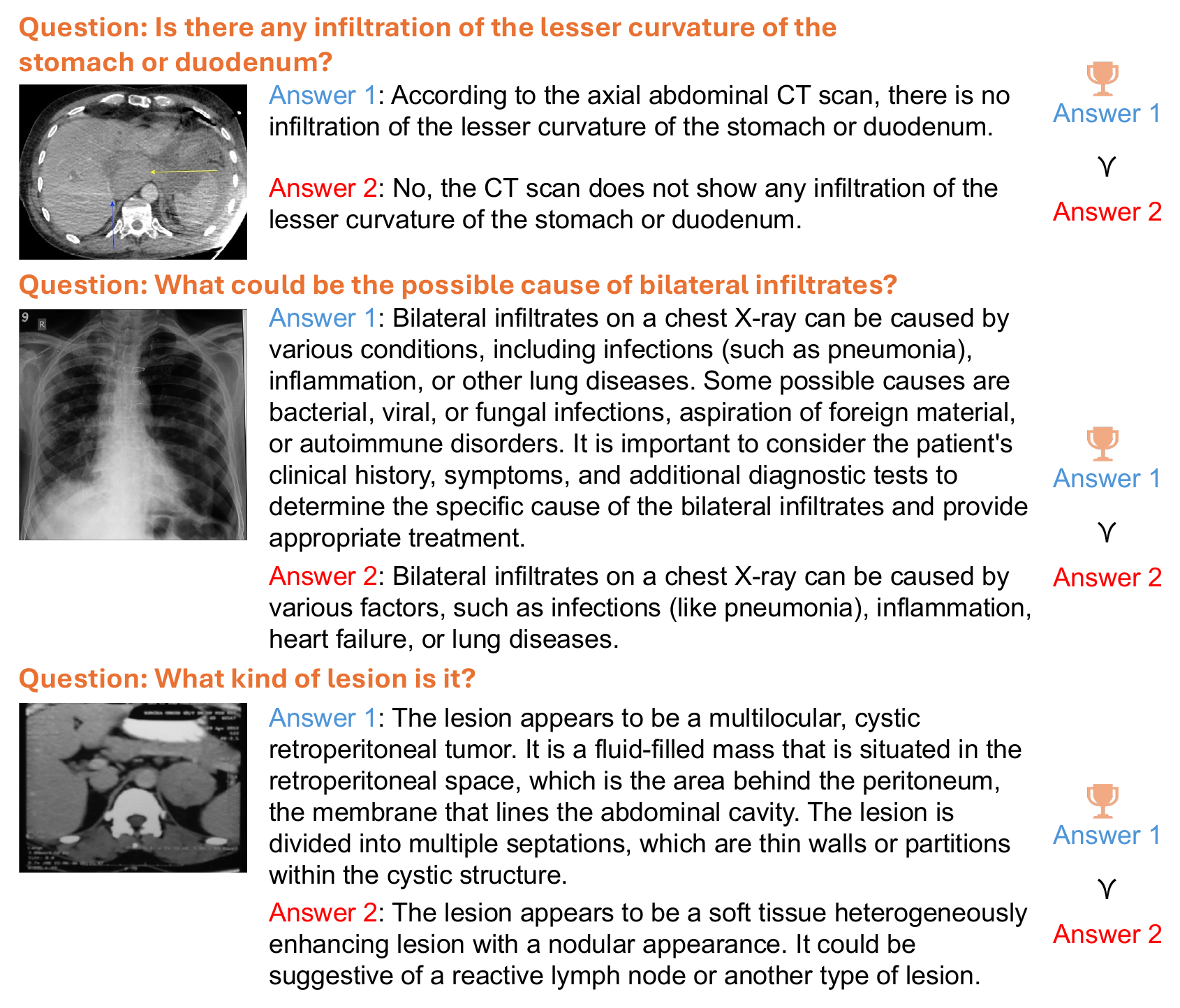}
\caption{ \label{fig:prefdata} Preference data visualization. The win and loss answer were classified by GPT-4o.}
\end{figure*}

\section{Experiments}

\subsection{Additional Results}\label{sec:moreresults}
In addition to evaluating zero-shot performance, we conducted experiments involving fine-tuning the model on downstream tasks. To maintain task generalizability and domain specificity, we compiled a new medical instructional dataset, combining Med-IM~\cite{li2023llavamed} with QA pairs from the training sets of VQA-RAD, SLAKE, and PVQA. Table~\ref{tab:extradata} details the number of medical images and QA pairs within each dataset.
After fine-tuning the models on this visual instruction dataset, we observed a clear improvement in downstream task performance, as shown in Table~\ref{tab:ft}. The performance gap between the baseline model LLaVA-v1.5 and the proposed STLLaVA-Med demonstrates the effectiveness of the self-training pipeline. 

Additionally, the improvement between STLLaVA-Med without DPO and STLLaVA-Med illustrates the effectiveness of preference alignment within the self-training pipeline. However, we found this improvement is not as significant as the improvement over zero-shot scenario. One explanation is the inconsistency between our designed preference and the ground truth preference. For VQA-RAD, SLAKE, and PVQA, the ground truth are short phrases, but the preference we are trying to optimize is \emph{detailed} and \emph{relevance}. This gives us an insight that the human expert should be involved in future medical tasks evaluation.

\subsection{Qualitative Results}\label{sec:apenqualiresult}
In Table~\ref{tab:main}, we have observed a clear improvement of model performance after preference optimization. In Fig.~\ref{fig:supsample}, we provide more qualitative results of medical VQA. As can be seen, STLLaVA-Med follows human preference by generating more detailed and accurate answers than the model without DPO fine-tuning.
Fig.~\ref{fig:prefdata} provides example samples of the preference data generated by STLLaVA-Med and GPT-4o. From these three samples, we find that the chosen answers contain more detail and in-depth analysis, aligning with human preference.

\end{document}